\documentclass{article}

\usepackage[accepted]{icml2024}

\usepackage{microtype}
\usepackage{graphicx}
\usepackage{subfigure}
\usepackage{booktabs} 

\usepackage{hyperref}

\usepackage{amsmath}
\usepackage{amssymb}
\usepackage{mathtools}
\usepackage{amsthm}

\usepackage[capitalize,noabbrev]{cleveref}

\theoremstyle{plain}
\newtheorem{theorem}{Theorem}[section]
\newtheorem{proposition}[theorem]{Proposition}

\theoremstyle{definition}

\theoremstyle{remark}

\usepackage{url}            
\usepackage{booktabs}       
\usepackage{amsfonts}       
\usepackage{amsmath}
\usepackage{amssymb}
\usepackage{nicefrac}       
\usepackage{microtype}      
\usepackage{xcolor}         
\usepackage{csquotes}

\usepackage{paralist, amsmath, amssymb, graphicx}
\usepackage{amssymb}
\usepackage{mathtools}
\usepackage{nccmath}
\usepackage{xcolor}

\usepackage{nicefrac}
\usepackage{stmaryrd}
\usepackage{bm}
\usepackage{cancel}
\usepackage{multirow}

\usepackage{xfrac}
\usepackage{mathrsfs}
\usepackage{amscd}
\usepackage{dsfont}
\usepackage{tikz}
\usepackage{epstopdf}
\usepackage{enumerate}
\usepackage{bm}
\usepackage{chngcntr} 
\usepackage{makecell}
\usepackage{placeins}
\usepackage{algorithm}
\usepackage{algorithmic}
\usepackage{multicol}
\usepackage{xargs}
\usepackage{soul}

\usepackage[textsize=tiny]{todonotes}

\newcommand{\la}{\lambda}
\newcommand{\sa}{\sigma}

\renewcommand{\l}{\left}
\renewcommand{\r}{\right}

\newcommand{\norm}[1]{\left\lVert#1\right\rVert}
\newcommand{\normbig}[1]{\big\lVert#1\big\rVert}
\newcommand{\abs}[1]{\l|#1\r|}

\newcommand{\red}{\color{black}}
\newcommand{\orange}{\color{black}}

\usepackage{comment}

\DeclareMathOperator{\prox}{prox}

\icmltitlerunning{}

\begin{document}


\twocolumn[
\icmltitle{A new computationally efficient algorithm to solve Feature Selection for Functional Data Classification in high-dimensional spaces}



\icmlsetsymbol{equal}{*}


\begin{center}
\large
    Tobia Boschi\textsuperscript{\rm 1, *},
    Francesca Bonin\textsuperscript{\rm 1},
    Rodrigo Ordonez-Hurtado\textsuperscript{\rm 1}, \\
    Alessandra Pascale\textsuperscript{\rm 1}, and 
    Jonathan Epperlein\textsuperscript{\rm 1} \\
    \vspace{0.1 cm}
    \textsuperscript{\rm 1}IBM Research Europe, Dublin \\
    \vspace{0.2cm}
\normalsize
    \textsuperscript{\rm *} Corresponding Author: tobia.boschi@ibm.com
\end{center}





\vskip 0.13in
]






\begin{abstract}

This paper introduces a novel methodology for Feature Selection for Functional Classification, FSFC, 
that addresses the challenge of jointly performing feature selection and classification of functional data in scenarios with categorical responses and multivariate longitudinal features. FSFC tackles a newly defined optimization problem that integrates logistic loss and functional features to identify the most crucial variables for classification.
To address the minimization procedure, we employ functional principal components and develop a new adaptive version of the Dual Augmented Lagrangian algorithm.
The computational efficiency of {FSFC} enables handling high-dimensional scenarios where the number of features may considerably exceed the number of statistical units.
Simulation experiments demonstrate that {FSFC} outperforms other machine learning and deep learning methods in computational time and classification accuracy. Furthermore, the {FSFC} feature selection capability can be leveraged to significantly reduce the problem's dimensionality and enhance the performances of other classification algorithms.   
The efficacy of {FSFC} is also demonstrated through a real data application, analyzing relationships between four chronic diseases and other health and demographic factors.
\end{abstract}

%
%
\section{Introduction}
\label{sec:intro}

Many contemporary scientific domains that require the integration and interpretation of time-dependent variables have recently benefited from Functional Data Analysis (FDA) \citep{ramsay2005, horvath2012inference}, an active domain within statistics and Machine Learning (ML) that allows working with multivariate longitudinal measurements by estimating smooth curves from multiple observations over a continuous domain. This functional representation facilitates the comparison of different statistical units and variables throughout the entire temporal domain.
FDA has demonstrated its effectiveness in many recent applications where the emergence of novel technologies, such as brain sensors, DNA sequencers, and wearable devices, has facilitated 
the measurement of a large number of longitudinal variables
\citep{shcherbina2017accuracy, cremona2019functional, smuck2021emerging}.
In these modern domain specific applications of FDA, two criticality emerge: \textit{1)} high dimensionality, the number of features may significantly exceed the number of statistical units \citep{ovsyannikova2020role, paul2014intervention}, \textit{2)} data scarcity \cite{9412492}. The first issue may lead ML and Deep Learning (DL) models to overfit, the second may lead current feature selection models to sub-optimal efficiency. \\
Hence, there is a growing demand for new algorithms able to 
i) effectively reduce problem dimensionality and feature complexity, ii) deal with data scarcity \cite{10.1016/j.knosys.2022.110111}, and iii) preserve computational efficiency \cite{9412492}.

To address these challenges in instances characterized by multivariate longitudinal variables and binary categorical responses, we introduce a new feature selection methodology called Feature Selection for Functional Classification (\texttt{FSFC}). 
The key innovative aspects of our work are the following: 



\vspace{-0.4cm}
\begin{list}{$\square$}{\leftmargin=1em \itemindent=0em}
    \item[\footnotesize $\bullet$] We define a new optimization problem that integrates logistic loss and multivariate functional features to concurrently solve feature selection and classification.

    \vspace{-0.2cm}
    \item[\footnotesize $\bullet$] We employ Functional Principal Components (FPC) and introduce an innovative adaptive version of the Dual Augmented Lagrangian (DAL) algorithm to address the optimization problem. FPC are used to simplify the complexity of each feature while retaining its longitudinal representation. On the other hand, DAL employs the sparsity structure of the problem to reduce its dimensionality. This combination of FPC and DAL makes \texttt{FSFC} highly computationally efficient, tackling high-dimensional scenarios where the number of features is much larger than the number of statistical units.

    \vspace{-0.2cm}
    \item[\footnotesize $\bullet$] We evaluate the efficacy of \texttt{FSFC} both in isolation and as a pre-processing step to enhance other ML and DL methods; we validate our approach through simulations and a real data application. In the real data application, we examine the relationships between four chronic diseases and other health and socio-demographic factors, demonstrating \texttt{FSFC} adaptability to multi-modal data scenarios.
\end{list}



\paragraph{Related Work.}
Feature selection has been a fertile field of
research and development since the 1970s and is now among the most renowned and widely utilized techniques in ML and DL to decrease the dimensionality of optimization problems, remove irrelevant and redundant features, improve efficiency in learning and predictive tasks, and enhance model interpretability \cite{blum1997selection, dash1997feature}. 
It has been applied to several tasks, including linear regression \citep{tibshirani1996regression, zou2005regularization}, and classification \citep{cai2018feature} for handling a vast number of instances and when dealing with high-dimensional data \cite{liu2002discretization, das2001filters}.
Additionally, feature selection has been extensively studied also in the context of linear functional regression  \citep{chen2016variable, parodi2018simultaneous, boschi2021highly}. 
However, while there are numerous existing methods for classifying longitudinal observations, including \emph{supporting vector machines (SVM)} \citep{muller1997predicting}, \emph{long short-term memory (LSTM) networks}  \citep{hochreiter1997long}, \emph{shapelets} \citep{ye2009time}, and \emph{functional classification} \citep{leng2006classification, fraiman2016feature}, to the best of our knowledge, no study to date has been conducted on approaches that enable simultaneous feature selection and classification within the framework of multivariate functional data. Indeed, although established techniques like \emph{Variational AutoEncoders} \citep{kingma2019introduction} and \emph{Time Series Feature Extraction} \citep{barandas2020tsfel} can reduce the complexity of each longitudinal feature by identifying key attributes, they are not designed to decrease the total number of features detecting the most relevant ones and do not preserve their longitudinal information.

In the remainder of this paper, we first present our methodology in Section \ref{sec:methodos}, we benchmark its performance against SVMs and LSTMs via simulations (Section \ref{sec:sim}) and a real-world application on the longitudinal SHARE dataset (Section \ref{sec:share}). We draw our conclusions in Section \ref{sec:discussion}.

%
%
\section{Methods}
\label{sec:methodos}


\subsection{Problem definition}

\texttt{FSFC} introduces a novel optimization problem that can be described as follows. 
Let the number of observations and features be denoted by $n$ and $p$, respectively. Let $\mathcal T = [a,b]$ be a closed bounded interval. Without loss of generality, we consider $\mathcal T$ = $[0,1]$.  The categorical responses are represented by $Y_i \in \{-1, 1\}$, the functional features by $\mathcal{X}_{ij} \in \mathbb{L}^2(\mathcal T)$, and the functional model coefficients by $\mathcal B_{j} \in \mathbb{L}^2(\mathcal T)$, for $i=1, \dots, n$ and $j=1, \dots, p$. It is assumed, without loss of generality, that $\mathcal{X}_{ij}$ is standardized with a mean function of 0 and standard deviation of 1. 
Finally, for a generic function $f \in \mathbb L^2(\mathcal{T})$ the squared $\mathbb{L}^2$-norm is defined as
$\lVert f\rVert_{\mathbb{L}^2}^2= \langle f, f\rangle_{\mathbb{L}^2} = \int_\mathcal{T} f^2(t) \text{dt}$.
The \texttt{FSFC} optimization problem is then formulated as follows:
%
\begin{align}
\small
\begin{split}
    \label{eq:functional_minimization_problem}
    \min_{\mathcal{B}_1, \dots, \mathcal{B}_p} 
    \Bigg[ 
    &\sum_{i=1}^n \log \bigg(1 + \exp \Big( -Y_i \sum_{j=1}^p \int_\mathcal{T}\mathcal{B}_j(t)\mathcal{X}_{ij}(t)dt \Big) \bigg) \\ 
    & + \sum_{j=1}^{p} \omega_j \left(
     \lambda_1 \lVert \mathcal{B}_j\rVert_{\mathbb{L}^2}+ 
     \frac{\lambda_2}{2}\lVert \mathcal{B}_j \rVert_{\mathbb{L}^2}^2 
    \right)  \Bigg]
\end{split}
\end{align}
The first term of the objective function \eqref{eq:functional_minimization_problem} is the logistic loss with functional features and enables classification. The second term comprises two different penalties that induce sparsity to facilitate feature selection. The parameters $\lambda_1$ and $\lambda_2$ control the global importance of the penalties with respect to the logistic loss. The feature-specific weights $\omega_j$ extend the \textit{adaptive} LASSO and Elastic Net \citep{zou2006adaptive, zou2009adaptive} to the functional classification settings and can improve feature selection by 
reducing the active set cardinality.

In order to solve \eqref{eq:functional_minimization_problem},  first, we approximate the functional variables 
via the Functional Principal Components (FPC) method \citep[see, e.g.][]{horvath2012inference}, and subsequently, perform the minimization process using a Dual Augmented Lagrangian (DAL) algorithm. 
The DAL algorithm was introduced \citet{tomioka2009dual, li2018} to solve LASSO and Elastic-Net problems and has been extended by \citet{boschi2023fasten} to accommodate functional linear regression cases. DAL leverages the intrinsic sparsity of the problem and the sparsity induced by the Hessian matrix information, resulting in a substantial reduction in computational cost. 
{\red In this work, we devise a novel adaptive version of DAL and develop the mathematical theory underpinning the algorithm to incorporate logistic loss and functional features within the objective function.}

\subsection{Matrix representation}
\label{subsec:matrix_representation}

To address \eqref{eq:functional_minimization_problem}, we employ a matrix representation obtained by expressing functional variables as linear combinations of basis functions \citep{ramsay2005, kokoszka2017introduction}. 
Specifically, we represent the functional features using their first $k$ FPCs. FPCs are a well-established technique to solve the functional regression \citep{reiss2007functional} and classification \citep{preda2007pls, wang2016functional} problems. We now show how to use FPCs to obtain a matrix representation of \eqref{eq:functional_minimization_problem}.

Consider
$\mathcal{X}_j = \left[\mathcal{X}_{1j} ~ | ~ \dots ~ | ~ \mathcal{X}_{nj} \right]^T$ representing the set of $n$ curves of the $j$-th feature. 
Let $e^j = \big[e^j_1 ~ | ~ \dots ~ | ~ e^j_k \big]$ denote the matrix containing the first $k$ FPCs of $\mathcal{X}_j$.
Define $X = \left[X_{[1]} | \dots | X_{[p]} \right] \in \mathbb{R}^{n \times pk}$ and $B = \left(B_{1}^T, \dots, B_{p}^T \right)^T \in \mathbb{R}^{pk}$, where $X_{[j]} \in \mathbb{R}^{n \times k}$ and $B_{j} \in \mathbb{R}^{k}$ are the score matrix of $\mathcal X_j$ and the score vector of $\mathcal B_j$ with respect to the same basis system $e^j$, respectively, for $j=1,\dots,p$. 
Specifically, $X_{[j](is)} = \langle \mathcal X_{ji}, e^j_s \rangle_{\mathbb{L}^2}$ and $B_{j,s} = \langle \mathcal B_{j}, e^j_s \rangle_{\mathbb{L}^2}$. 
Then, we can approximate $\mathcal X_j(t) \approx X_{[j]}e^j(t)^T, ~ \mathcal B_j(t) \approx e^j(t)B_{j}$, and  $\int_\mathcal{T}\mathcal{B}_j(t)\mathcal{X}_{ij}(t)dt \approx X_{[j](i)}\left(\int e^j(t)^Te^j(t)dt \right) B_{j}$. 
Note that we use the notation $B_{j,s}$ and $X_{[j](i)}$ to indicate the $s$-th element of the vector  $B_{j}$ and $i$-row of the matrix $X_{[j]}$, respectively.
Since $e^j$ are orthonormal bases, then $\int e(t)^Te(t)dt = I_k$, i.e., the identity matrix of order $k$.
Moreover, for a generic function $f \in \mathbb L^2([0,1])$, one has that $\lVert f \rVert_{\mathbb{L}^2}^2 = \sum_{i=1}^\infty \langle f, e_i\rangle_{\mathbb{L}^2}$ \citep{kokoszka2017introduction}. Thus, we can approximate the ${\mathbb{L}^2}$ functional norm and the standard $l_2$ vector norm, i.e., the Frobenius norm, denoted by $\lVert \cdot \rVert_2$. 
We can now express \eqref{eq:functional_minimization_problem} in a matrix form as:
\begin{align}
\label{eq:matrix_minimization_problem}
\small
\begin{split}
	     \min_{B} ~ 
     \Bigg[
     \sum_{i=1}^n \log \bigg(1 + \exp \Big( -Y_i \cdot \big(X_{(i)}B \big) \Big) \bigg) \\ 
     + \sum_{j=1}^{p} \omega_j \left( \lambda_1 \lVert B_{j}\rVert_{2}+\frac{\lambda_2}{2} \lVert B_{j}\rVert_{2}^2 \right)\Bigg]  \ ,
\end{split}
\end{align}
where $X_{(i)}$ indicates the $i$-th row of $X$. 
Mimicking the Group Elastic Net regularization \citep{zou2005regularization, simon2013sparse}, Equation \eqref{eq:matrix_minimization_problem} combines two penalties: the first is non-differentiable, creating sparsity, while the second (Ridge-like) is differentiable, controlling multicollinearity and accelerating the convergence of the optimization algorithm.
Once we obtain an estimate of the score vector $\hat B$, the coefficient curves can be recovered as
$\hat{\mathcal B}_j(t) = e^j(t)\hat B_j(t)$, the class probability as 
$\hat p_i = 1/\big(1 +  \exp \big(X_{(i)} \hat B \big) \big)$, 
and the categorical response as $\hat Y_i = 1$ if $\hat p_i > 0.5$ and $-1$ otherwise.

\paragraph{Selection of K.} 
The selection of $k$ determines the extent to which Equation \eqref{eq:matrix_minimization_problem} approximates Equation \eqref{eq:functional_minimization_problem}. 
FPCs have the significant advantage of being the most parsimonious orthonormal basis system: in most instances, a few components capture more than 90\% of the curves' variability. As discussed in the next section, using a small $k$ is critical for the efficiency of the DAL algorithm. The choice of $k$ also accounts
for the trade-off between computational efficiency and approximation accuracy.
In the current version of \texttt{FSFC}, we require $k$ to be identical across all features, i.e., one must use the same number of basis components for all features. In the Discussion section, we further detail the possibility of accommodating a distinct $k$ for each feature.

\subsection{Dual Augmented Lagrangian (DAL) Algorithm}
\label{subsec:dual}

The core idea underlying the DAL methodology involves minimizing the \emph{Augmented Lagrangian} function associated with the \emph{dual} problem. Prior to defining the dual problem, note that \eqref{eq:matrix_minimization_problem} can be expressed as
\begin{equation}
    \label{eq:primal}
    \tag{P}
    \min_B \big[~h(XB) + \pi(B) ~\big],
\end{equation}
where 
$h(XB)= \sum_{i=1}^n \log \big(1 + \exp \big( -Y_i \cdot (X_{(i)}B ) \big) \big)$ 
is the logistic loss function and $\pi(B) = \sum_{j=1}^{p} \omega_j ( \lambda_1 \lVert B_{[j]}\rVert_{2} +\frac{\lambda_2}{2} \lVert B_{[j]}\rVert_{2}^2)$ is the adaptive Elastic Net-type penalty. 
Note that the penalty $\pi(B)$ is the same as that employed by \citet{boschi2023fasten} in the scalar-on-function regression scenario. Although their results can be utilized for penalty-related operators, a novel theoretical framework must be developed for the functional logistic loss. 
{\red In particular, one needs to redefine a set of mathematical operators in broader dimensions while maintaining the DAL sparsity structure and efficiency.}

\paragraph{Dual problem.} 
A potential dual representation of the primal \eqref{eq:primal} can be derived from \citep{boyd2004convex, tomioka2011super} as:
%
\begin{equation}
    \label{eq:dual}
    \tag{D}
    \min_{V,Z} \big[~h^*(V) + \pi^*(Z)~\big]\quad s.t.\quad X^TV+Z=0 \ . 
\end{equation}
In this formulation, $V\in\mathbb{R}^{n}$ and $Z\in\mathbb{R}^{pk}$ are the dual variables. Following the notation introduced for $B$, we can express $Z = \left(Z_{1}^T, \dots, Z_{p}^T \right)^T$, where $Z_{j}\in \mathbb{R}^{k}$ denotes the sub-vector of $Z$ corresponding to the $j$-th feature.
$h^*$ and $\pi^*$ denote the Fenchel-conjugate functions \citep{fenchel1949conjugate} of $h$ and $\pi$, respectively. 
The computation of $\pi^*$ is derived in \cite{boschi2023fasten} as
\begin{align}
\small
	\begin{split}
    \label{eq:p_star}
        	\pi^*(Z) = \sum_{j=1}^p \pi^*(Z_{j}) = \sum_{j=1}^p (2\omega_j\la_2)^{-1} \left(\big[\norm{Z_{j}}_2 - \omega_j\la_1\big]_{+}\right)^2 ,
    \end{split}
\end{align}
%
where $[~\cdot~]_{+}$ is the positive part operator: $[s]_{+} = s$ if $s>0$, and $0$ otherwise. 
The form of $h^*$ is given in the next Proposition (a proof can be found in the Appendix, Section \ref{sec:appendix_proofs}).
%
\begin{proposition}
Considering $h$ as in Equation \eqref{eq:primal}, then the function $h^*$ is defined for $\abs{Y_i V_i} < 1$ as follows:
\begin{equation}
    \small
    	\label{eq:h_star}
        	h^*(V) = \sum_{i=1}^n (1 - \abs{Y_i V_i})\log (1 - \abs{Y_i V_i}) + \abs{Y_i V_i}\log (\abs{Y_i V_i}) \ .
\end{equation}
\end{proposition}
\noindent
We are now ready to define the \textit{Dual Augmented Lagrangian} function as
%
\begin{equation}
\label{eq:augmented_lagrangian}
\small
\begin{split}
    \mathcal{L}&_\sigma (V,Z,B) =   
    h^*(V) + \pi^*(Z) \\
     &- \sum_{j=1}^p \langle B_{j}, V^TX_{j} + Z_{j} \rangle + \frac{\sigma}{2} \sum_{j=1}^p \lVert V^TX_{j} + Z_{j}\rVert_2^2 \ ,
\end{split}
\end{equation}
with $\sigma>0$.
Note that the Dual Augmented Lagrangian is defined by augmenting the dual objective function with a penalization term that accounts for the violation of the problem constraint.
%
%
\begin{algorithm}[t]
\caption{\textbf{DAL Method}}
\label{alg:dal}
\footnotesize
\begin{algorithmic}
\vspace{0.2cm}
\STATE \textbf{GOAL:} minimize  $\mathcal{L}_\sigma (V,Z,B)$. Start from the initial values $V^0, Z^0, B^0, \sa^0$ \\
\vspace{0.2cm}
\STATE \texttt{WHILE NOT CONVERGED:}
\begin{itemize}
    
    \vspace{-0.1cm}
    \item[\textbf{(1)}] Given $B^s$, find $V^{s+1}$ and $Z^{s+1}$ which approximately solve:
    \vspace{-0.3cm}
    \begin{equation}
        \label{eq:inner_subproblem}
         	 \big(V^{s+1}, Z^{s+1}\big) \approx \arg \min_{V,Z} \mathcal{L}_\sa \big(V, Z ~|~ B^s\big)
    \end{equation}
    \vspace{-0.3cm}
    
    \fbox{\begin{minipage}{22.5em}
    
        \vspace{0.1cm}
        ~ ~\textbf{Inner sub-problem:} to find $(V^{s+1}$, $Z^{s+1})$, \\
        $~ ~ ~$ update V and Z \emph{independently}: \\
        
        \vspace{-0.2cm}
        ~ ~ ~ ~\texttt{WHILE NOT CONVERGED}   
        
        \begin{itemize}
            
            \vspace{0.1cm}
            \item[] ~ ~ ~$V^{m+1} = \arg\min_V \mathcal{L}_\sigma (V\,|\,Z^m,B^s)$ \\
            $~ ~ ~ ~ ~ ~ ~ ~ ~\longrightarrow~$ {\footnotesize \textbf{Newton method} - see Proposition (2)}
            
            \vspace{0.1cm}
            \item[] ~ ~ ~$Z^{m+1} = \arg\min_Z \mathcal{L}_\sigma (Z,|\,V^{m+1},B^s)$ \\
            $~ ~ ~ ~ ~ ~ ~ ~ ~\longrightarrow~$ {\footnotesize \textbf{closed-form} - see \eqref{eq:z_bar}}
            \vspace{0.1cm}
        
        \end{itemize}
    
    \end{minipage}}
    
    \vspace{0.1cm}
    \item[\textbf{(2)}] Update the Lagrangian multiplier $B$ and the parameter $\sa$:
    
    \vspace*{-0.2cm}
    \begin{equation*}
    \label{eq:update_x}
    \begin{split}
                        &B^{s+1} = B^s - \sa^s \big(X^T V^{s+1} + Z^{s+1}\big), \ 
                        \sa^{s} \le \sa^{s+1} < \infty
    \end{split}
    \end{equation*}
\end{itemize}
\vspace*{-0.3cm}
\end{algorithmic}
\end{algorithm}
Our DAL implementation is outlined in Algorithm \ref{alg:dal}, as adapted from \citet{boschi2023fasten}.
The core part of the algorithm involves solving the inner sub-problem \eqref{eq:inner_subproblem}, which entails minimizing the Dual Augmented Lagrangian with respect to the dual variables $V$ and $Z$. To obtain an approximate solution of \eqref{eq:inner_subproblem}, $V$ and $Z$ can be updated independently \citep{tomioka2009dual}.
The primal variable $B$ and the parameter $\sigma$ are updated according to standard rules \citep{li2018}. The choice of $\sa^0$ is empirically studied in \citet{tomioka2011super}: {\red starting from very small values of $\sa$ results in an increased number of iterations required for convergence; however, if $\sa^0$ is excessively large or increases too rapidly, the DAL method fails to converge to the optimal solution.}

\paragraph{Update of Z.}
The minimization with respect to $Z$ has a closed-form solution $\bar Z$, which is computed in \citet{boschi2023fasten} based on the \emph{proximal operator} \citep{rockafellar1976augmented,rockafellar1976} of $\pi$, denoted as $\prox_{\sigma \pi}$. Specifically, the following holds:
\begin{align}
\small
\label{eq:z_bar}
	\Bar{Z} &= \prox_{\pi^*/\sa} \left(B/\sa - X^T\Bar{V} \right) \nonumber 
	\\
	&= B/\sa - X^T\Bar{V} - \prox_{\sa p}\left(B - \sa X^T\Bar{V}\right)/\sa \ ,
\end{align}
where $\prox_{\sigma \pi}(B) = \big(\prox_{\sigma \pi}(B_{1}), \dots, \prox_{\sigma \pi}(B_{p}) \big)^T$, and for each $j$
\begin{equation}
\small
\label{eq:prox_op}
	\prox_{\sigma \pi}(B_{j}) =  (1 + \sa \omega_j \la_2)^{-1} \left[1- \norm{B_{j}}_2^{-1} \sa \omega_j \la_1  \right]_{+} B_{j} \ .
\end{equation}
%

\paragraph{Update of V.} Given that the minimization with respect to $V$ lacks a closed-form solution, we update $V$ applying the \emph{Newton Method} \citep{nocedal1999numerical} to the function $\psi(V) := \mathcal{L}_\sigma (V\,|\,\Bar{Z},B)$.
The Newton update is expressed by $V^{m+1} = V^{m} + sD$, where $D \in \mathbb{R}^{n}$ represents the descent direction and $s$ the step-size. 
To obtain $D$, we solve the linear system
\begin{equation}
\label{eq:newton_direction}
    H_{\psi}(V) D = -\nabla \psi (V) \ , 
\end{equation}
%
where $H_{\psi} \in \mathbb R^{n \times n}$ and $\nabla \psi \in \mathbb R^{n}$ are the Hessian matrix and the gradient vector of $\psi$, respectively.
To determine the step size, we implement the \emph{line-search} strategy proposed by \citet{li2018}. Starting with $s=1$, we iteratively reduce it by a factor of $0.5$ until the condition
\begin{equation}
\label{eq:linesearch}
	\psi \big(V^{m+1} \big) \le \psi (V^m) + \mu s\nabla \psi^T (V^m) D
\end{equation}
is met, with $\mu \in (0, 0.5)$. 
Solving the linear system \eqref{eq:newton_direction} is the most computationally demanding step in the DAL algorithm.  
However, the following proposition leverages the sparsity structure of $H_{\psi}$ to significantly reduce the system dimension and computational complexity (a proof can be found in the Appendix, Section \ref{sec:appendix_proofs}).
\begin{proposition}
Let $T=B-\sa X^T V$, $T_{j}=B_{j} - X_{[j]}^T V$, $\mathcal{J} = \big\{ j~:~ \lVert T_{j} \rVert_2  \ge \sa \omega_j \la_1 \big\}$, and $r = |\mathcal{J}|$ be the cardinality of $\mathcal{J}$.
Next, let $ X_\mathcal{J} \in \mathbb{R}^{n \times rk}$ be the sub-matrix of $X$ restricted to the blocks $X_{j}$, 
$j \in \mathcal{J}$.
Define the square $k \times k$ matrix $P_{[j]}$ as
\begin{equation*}
    \small
    \begin{split}
        P_{[j]} = (1 + \sa \omega_j \la_2)^{-1} \left( \left(1 -
        \frac{\sa \omega_j \la_1}{ \norm{T_{j}}_2} \right) I_{k} +
        \frac{\sa \omega_j \la_1}{ \norm{T_{j}}_2^3} 
        T_{j}T_{j}^T \right) \ .
    \end{split}
\end{equation*}
Finally, let $Q_\mathcal{J} \in \mathbb{R}^{rk \times rk}$ be the block-diagonal matrix formed by the blocks $P_{[j]}$, $j \in \mathcal{J}$.
Then:
  \begin{align*}
    \small
    \begin{split}
        &(i)~ \  \ \psi(V) = h^*(V)\\
        & ~ ~ ~ ~ ~ ~ ~ ~ ~ ~ ~ ~ ~ + \frac{1}{2\sa}  \sum_{j=1}^p 
        \left( \left( 1 + \sa \omega_j \la_2 \right) \normbig{\prox_{\sa \pi}\big(T_{j}\big) }_2^2 - \norm{B_{j}}_2^2 \right) \ , \\ 
        &(ii)~ \ \nabla \psi(V) = \nabla h^*(V) - X \prox_{\sa \pi}(T) \ , \\
        &(iii) ~H_{\psi}(V) = H_{h^*}(V)  + \sa X_\mathcal{J} Q_\mathcal{J} X_\mathcal{J}^T \ ,
    \end{split}
    \end{align*}
where $\nabla h^*(V) \in \mathbb R^{n}$ and $H_{h^*}(V) \in \mathbb R^{n \times n}$ are the gradient vector and the Hessian matrix of $h^*$, respectively. Specifically, when $\abs{Y_iV_i} < 1$,
each element $i$ of $\nabla h^*(V)$ is equal to
$$
    Y_i \log \big((1 - \abs{Y_iV_i}) ^{-1} \abs{Y_iV_i}  \big) ~ ,
$$ 
and $H_{h^*}(V)$ is a diagonal matrix with elements given by
$$
    \big( \abs{Y_iV_i}(1 - \abs{Y_iV_i})\big)^{-1}.
$$
\end{proposition}

\paragraph{Computational efficiency.}
As stated in Proposition 2, the DAL algorithm takes advantage of the sparsity information embedded within the Hessian matrix, selecting a subset $\mathcal{J}$ of $r$ active features, with $r$ decreasing at each iteration.
By selecting an $r$-feature subset, the overall computational expense of solving the linear system \eqref{eq:newton_direction} is reduced from  $\mathcal{O} \l( n(n^2 + npk + p^2k^2) \r)$ to $\mathcal{O} \l( n(n^2 + nrk + r^2k^2) \r)$ with $r<p$.
In sparse scenarios, where the number of active features is low, $r$ might be significantly smaller than $n$, further reducing the computational burden through the application of the \emph{Sherman-Morrison-Woodbury} formula \citep{van1983matrix}: 
$\left(H_{h^*} + \sa X_\mathcal{J} Q_\mathcal{J} X_\mathcal{J}^T \right)^{-1} = 
H_{h^*}^{-1} -  H_{h^*}^{-1}X_\mathcal{J}  \big( \l( \sa Q_\mathcal{J} \r)^{-1} + X_\mathcal{J}^T H_{h^*}^{-1}X_\mathcal{J} \big)^{-1} X_\mathcal{J}^TH_{h^*}^{-1}$.
This equivalence enables the factorization of an $rk \times rk$ matrix, resulting in a total cost of $\mathcal{O} \l( rk(k^2 + nrk + r^2k^2 + n^2) \r)$ since $H_{h^*}$ is diagonal and computing its inverse is straightforward. 
Remarkably, the computational burden is not dependent on the total number of features $p$, but solely on the number of active features $r$. 
Provided that sparsity is maintained (i.e., $r$ remains small), the 
number of features can grow without impacting the efficiency of the linear system resolution. 
{\red However, it is important to acknowledge that a larger $p$ increases the cost of the principal components computation during the matrix representation stage, since the FPC scores have to be derived for each feature.}
Note that the computational cost depends on $k^3$. By maintaining a small value for $k$, which describes the number of bases used to represent the functional features,  the efficiency of the method is significantly enhanced.

\paragraph{Convergence criteria.} 
In order to guarantee the DAL super-linear convergence rate \citep{tomioka2009dual, li2018}, we implement consistent stopping-criteria to evaluate the convergence of 
both the inner sub-problem and the overall algorithm. With respect to the inner sub-problem, $Z$ and $V$ are updated iteratively until the following condition is satisfied \citep{tomioka2011super}:
\[
    \small
    \normbig{\nabla \psi\big(V^s\big)}_2 \le 2 \sa^{1/2} \sum_{j=1}^p \normbig{\big(X^T V^{s+1} + Z^{s+1}\big)_j}_2 ~ .
\]
For the overall algorithm's convergence, we monitor one of the \textit{Karush-Kuhn-Tucker} (KKT) conditions associated with the dual problem~\eqref{eq:dual}, namely, $X^T V = -Z$. 
This condition is exclusively satisfied  by the optimal solutions of \eqref{eq:dual} \citep{boyd2004convex}. Then, the algorithm is halted when the standardized residual of the KKT equation is less than a specified tolerance $tol$, as follows:
\begin{equation}
\label{eq:tol}
\small
    \Big(1 + \norm{V}_2 + \sum_{j=1}^p \norm{Z_j}_2 \Big)^{-1}
    \sum_{j=1}^p \norm{(X^T V + Z)_{j}}_2 < \text{tol} \ . 
\end{equation}


\subsection{Model selection and adaptive implementation}
\label{subsec:adaptive}


To assess the solution of \eqref{eq:matrix_minimization_problem} with varying values of the penalty parameter $\lambda_1$, we implement a path search mechanism.
%
%
We perform the search for $\lambda_1$ using the formula $\lambda_1 = c_\lambda\lambda_{max}$, with $c_\lambda$ belonging to a grid of $100$ values evenly spaced on a logarithmic scale from $1$ to $0.01$, and $\lambda_{max} = 0.5\max_j\lVert (X^TY)_{(j)}/\omega_j\rVert$. For $c_\lambda=1$ (i.e., $\lambda_1=\lambda_{max}$), $0$ active features are selected, and as $\lambda_1$ decreases for further values of $c_\lambda$, the number of features selected by the solution increases.
At this stage, the weights $w_j$'s in \eqref{eq:matrix_minimization_problem} are all equal to $1$.
We select $\lambda_1$ along the path that minimizes a \textit{5-fold cross-validation} classification accuracy score, denoting the optimal value of $\lambda_1$ and the corresponding solution as $\tilde{\lambda}_1$ and $\tilde{B}$, respectively.
The \emph{adaptive solution} is then computed starting from $\tilde{B}$.
%
We set $\omega_j=\text{sd}_B/\lVert \tilde{B}_{j}\rVert_2$, where $\text{sd}_B$ is the standard deviation of  $\big( \lVert\ \tilde{B}_1\rVert_2, \dots , \lVert \tilde{B}_r\rVert_2 \big)$, and we execute a single DAL minimization considering $\lambda_1 = \tilde{\lambda}_1$.
The adaptive feature-specific weights $w_j$'s allow for imposing a greater penalty on the coefficient curves that, despite a small norm, have not been screened out by the unweighted minimization, thus promoting their removal from the active set.

%
%
\section{Simulation study} 
\label{sec:sim}

\paragraph{Settings.} In this section, we evaluate the performance of \texttt{FSFC} using synthetic data. We benchmark \texttt{FSFC} against two representative techniques for longitudinal data classification: a \texttt{kernlab R package} \citep{kernlab} implementation of 
\texttt{SVM}, and a \texttt{TensorFlow} \citep{tensorflow2015} implementation of \texttt{LSTM}.
Furthermore, we apply \texttt{SVM} and \texttt{LSTM} on a reduced streamlined problem that considers solely the active features identified by \texttt{FSFC} and their FPC representation -- essentially, employing the \texttt{FSFC} output as input. 
{\red Notably, \texttt{SVM} and \texttt{LSTM} do not have any selection or problem-reduction capabilities.}
These approaches are called \texttt{r-SVM} and \texttt{r-LSTM}, respectively.
The hyper-parameters used for all the methods are discussed and detailed in the Appendix, Section \ref{sec:appendix_hyperparameters}. 

We consider two distinct scenarios: one with $n = 300$ and $p=800$, and another with $n = 600$ and $p=2000$. We denote the number of active features (i.e.,~non-zero regression coefficient curves) as $p_0$. For each scenario, we examine four different levels of sparsity by setting $p_0 = 2, 5, 10, 20$. 
{\orange The synthetic data are generated as described in Appendix Section \ref{sec:appendix_synthetic_data}.}

We assess 
\texttt{FSFC} in terms of \emph{selection} performance, while all the methods are evaluated with respect to \emph{classification accuracy} and \emph{computational efficiency}. 
To assess selection performance, we compute the \emph{recall} and \emph{precision} scores.
Classification accuracy is computed as the proportion of observation correctly classified, in both the \emph{training} and the \emph{test} sets, where the latter is generated independently from the former. We set the size of the test set $n_{test} = n/3$, with $n$ being the size of the training set.

\begin{figure*}[h!] 
    \centering
    \includegraphics[width=1\textwidth]{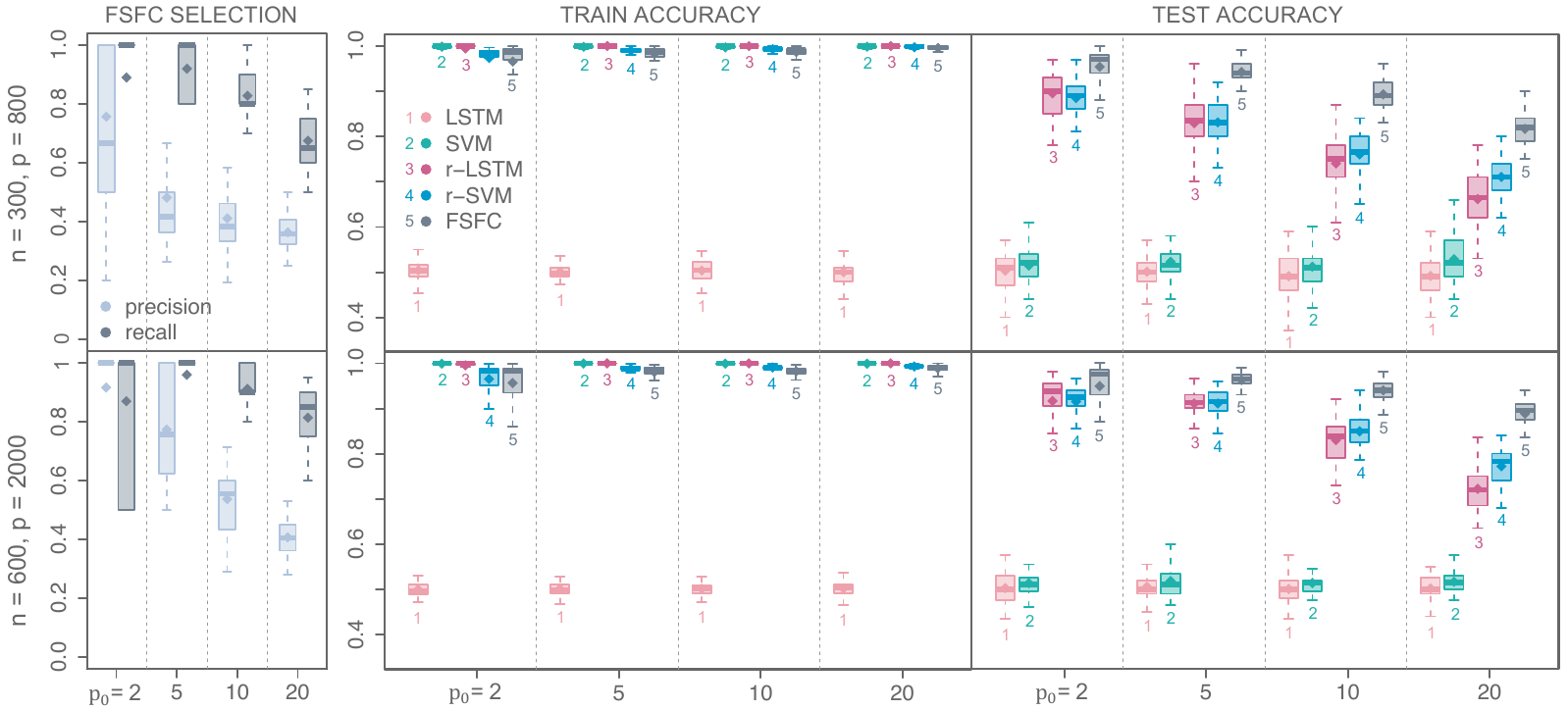}
    \vspace*{-0.5cm}
    \caption{Simulation results. Boxplots generated from the distribution obtained across 50 replications of each scenario, with gray diamonds and horizontal lines indicating means and medians of the distributions, respectively. Selection performances (precision and recall) are computed just for \texttt{FSFC}, while classification accuracy in the training/test set is reported for all the examined algorithms (\texttt{LSTM, SVM, r-LSTM, r-SVM, FSFC}). The rows illustrate two distinct scenarios ($n=300$, $p=800$, and $n=600$, $p=2000$). In each scenario, we investigate $p_0 = 2, 5, 10, 20$ (x-axes).
    }
\label{fig:sim_results}
\end{figure*}


\begin{table}[]
\caption{Simulation results. Average CPU processing time, measured in seconds, for 50 instances in each scenario. {\red All the computations were executed on a MacBookPro 2021 with an M1 Max processor and 32GB of RAM.}
}
\vspace{0.2cm}
\centerline{
\scalebox{0.88}{
	\begin{tabular}{l|r|rrr|rr}

                                                                         & $p_0$ & \texttt{LSTM}   & \texttt{SVM}   & \texttt{FSFC} & \texttt{rLSTM} & \texttt{rSVM}  \\ 
                                                                         \Xhline{2\arrayrulewidth}

\multirow{4}{*}{\begin{tabular}[c]{@{}l@{}} $n=300$ \\ $p=800$ \end{tabular}}   & \texttt{2}    & 140.61 & 16.17  
 & 1.44 & 5.56   & 0.01   \\
                                                                         & \texttt{5}    & 141.25 & 16.19  & 1.72  & 5.61   & 0.01  \\
                                                                         & \texttt{10}   & 139.93 & 16.17 & 1.84  & 5.66   & 0.02   \\
                                                                         & \texttt{20}   & 140.05 & 16.12  & 2.11 & 5.72   & 0.04  
                                                                         \\ \hline 
                                                                         \hline
                                                                        
\multirow{4}{*}{\begin{tabular}[c]{@{}l@{}} $n=600$ \\ $p=2000$\end{tabular}} & \texttt{2}    & 355.66 & 144.18 & 5.44  & 9.52   & 0.01  \\
                                                                         & \texttt{5}    & 354.16 & 142.14 & 7.09 & 9.65   & 0.02   \\
                                                                         & \texttt{10}   & 348.51 & 141.96 & 7.58 & 9.86   & 0.05   \\
                                                                         & \texttt{20}   & 349.58 & 142.82  & 8.16 & 10.11  & 0.13 \\ 
                                                                         \Xhline{2\arrayrulewidth}

\end{tabular}}}
\label{tab:sim1_time} 
\end{table}

\paragraph{Results.} The simulation study results are presented in Figure \ref{fig:sim_results}.
The displayed boxplots are derived from the distribution of the various scores obtained across 50 replications of each scenario.
In terms of \texttt{FSFC} feature selection performance, the recall score consistently surpasses the precision score, maintaining an average above $80\%$, except for the scenario with $n=600, p=2000$, and $p_0=20$. This suggests that while \texttt{FSFC} might occasionally select non-relevant features, it is proficient in identifying the active ones. Both metrics show a decline as the number of active features grows, indicating challenges in more complex scenarios.  

Except for \texttt{LSTM}, all models exhibit high classification accuracy in the training set.  \texttt{FSFC} outperforms all competitors in the test set, with an average accuracy larger than $85$\% when $p_0=2,5,10$. As the proportion of active features increases, i.e., in scenarios characterized by larger $p_0$ and smaller $p$, distinguishing non-relevant information becomes more challenging also for \texttt{FSFC}.
Notably, \texttt{r-SVM} and \texttt{r-LSTM} test accuracy outperforms the standard \texttt{SVM} and \texttt{LSTM} (especially in the scenario with a larger $p$): employing \texttt{FSFC} for a preliminary feature screening and problem dimension reduction results in a significant performance enhancement, potentially mitigating overfitting. 

The average CPU time reported in Table \ref{tab:sim1_time} demonstrates that \texttt{FSFC} is significantly more efficient than \texttt{SVM} and \texttt{LSTM}. This difference is even larger when $p$ increases. As expected, \texttt{FSFC}'s computational cost does not depend on $p$, but only on the number of active features $p_0$.
{\red By exploiting the properties of DAL optimization, \texttt{FSFC} can solve a problem with a large number of features ($p=2000$) in under $10$ seconds. Given that a $5$-fold cross-validation is implemented on a grid of 100 values, this implies that solving DAL takes, on average, less than $0.2$ seconds per repetition.}
Moreover, leveraging \texttt{FSFC} as a preliminary reduction step significantly trims the CPU time for both \texttt{r-SVM} and \texttt{r-LSTM}, given the much smaller feature set they operate upon.

{\orange Finally, in Appendix Table \ref{tab:fe_sim_results}, we report the accuracy and the CPU time related to a hybrid approach that combines Feature Extraction \citep{barandas2020tsfel} with \texttt{LSTM} and \texttt{SVM}. This approach does not reduce the number of features but simplifies the longitudinal variables by extracting salient attributes before performing the classification algorithms. Contrary to \texttt{rLSTM} and \texttt{rSVM}, results indicate that Feature Extraction does not enhance the accuracy of \texttt{LSTM} and \texttt{SVM} while significantly increasing the computation time.}

%
%
\section{SHARE application} 
\label{sec:share}


\begin{figure*}[!t]
    \centering
    \includegraphics[width=1\textwidth]{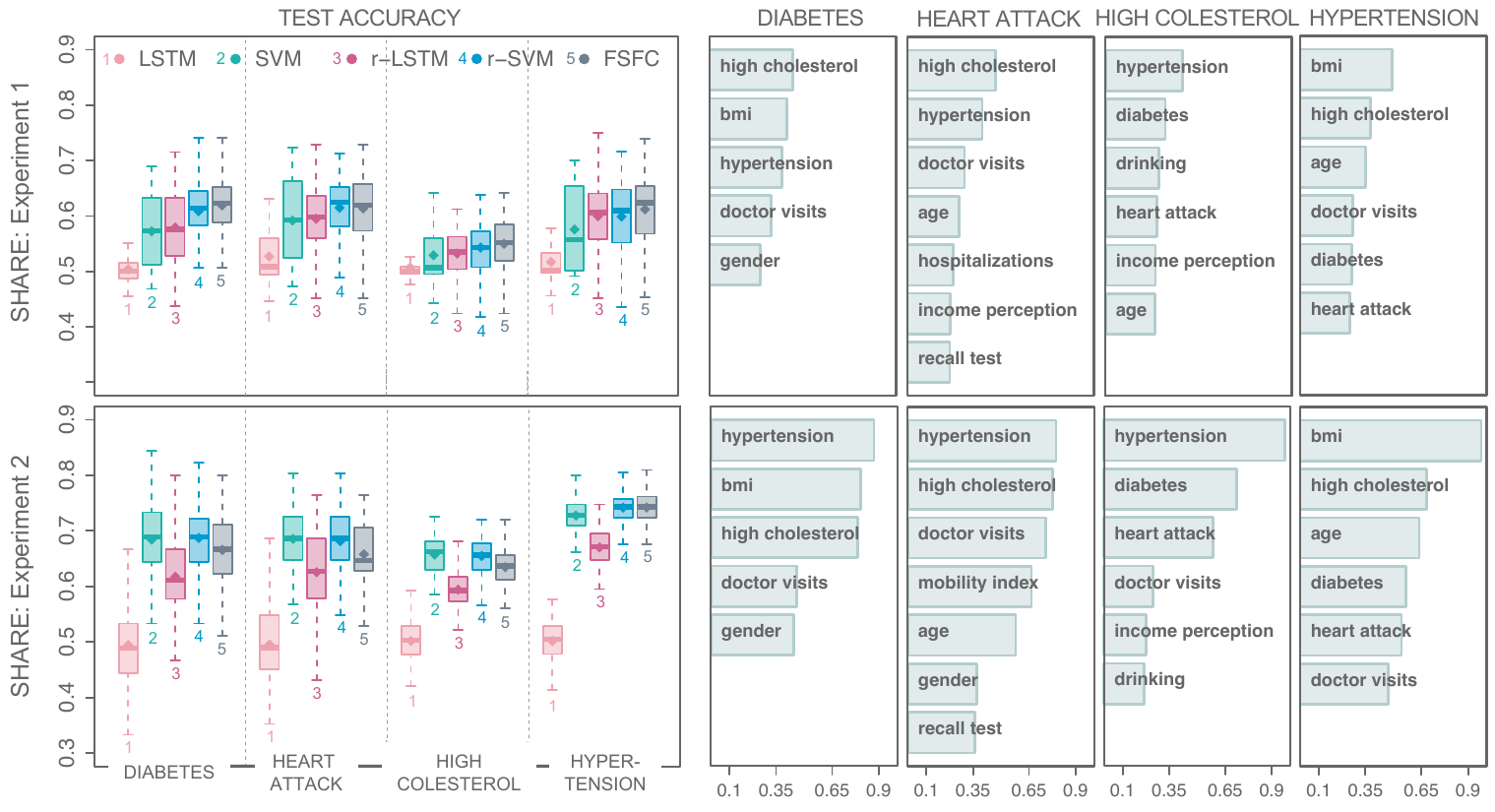}
    \vspace*{-0.5cm}
    \caption{Experiment 1 (upper panel) and Experiment 2 (lower panel) SHARE results. The test set classification accuracy boxplots (on the left) are generated from 100 replications. The dots and the horizontal lines indicate the means and medians of the distributions, respectively. 
    On the right, features selected by \texttt{FSFC} for more than 80 out of 100 replications are displayed for each response variable. The bar plots illustrate the average ratio of $\lambda_{max}$ at which each feature entered the active set. The higher the ratio, the earlier the feature is included in the model during the $\lambda$ path search.
        }
\label{fig:share_results}
\end{figure*}

The Survey of Health, Ageing and Retirement in Europe - SHARE \citep{mannheim2005survey} is a research infrastructure that aims to investigate the effects of health, social, economic, and environmental policies on the life course of European citizens \citep{borsch2013data, bergmann2017survey, borsch2020survey}.
SHARE is a longitudinal study, where the same subjects are followed over multiple years. Specifically, eight surveys or ``waves'' were conducted from 2004 to 2020 (see Appendix Figure \ref{fig:appendix} for the SHARE timeline).
We focus on the 1262 subjects who participated in at least seven out of the eight waves, ensuring a sufficient number of measurements for reliable curve estimation.
We investigate a subset of $p=36$ variables from the EasySHARE dataset \citep{gruber2014generating}, a preprocessed version of the SHARE data.
In our application, we aim to study the relationship between $4$ chronic diseases -- \emph{i.e., diabetes, myocardial infarction, high cholesterol}, and \emph{hypertension} -- and various physical and mental health, socio-demographic, and healthcare factors (a comprehensive list of the examined variables can be found in Appendix Table \ref{tab:var_list}).
While some of them are described by values that change over time (e.g., CASP index, max grip) and are suitable for a functional representation, others are scalar (e.g., education years) or categorical (e.g., gender) and do not evolve over different waves. 
We smoothed time-varying variables using cubic \emph{B-splines} with knots at each survey date and roughness penalty on the curve's second derivative \citep{ramsay2005}. For each curve, the smoothing parameter was selected by minimizing the average generalized cross-validation error \citep{craven1978smoothing}. 
{\red Although the surveys' dates and the amount of measurements may vary across subjects, the functional representation provides a natural imputation for missing values and facilitates the comparison of different statistical units across the entire temporal domain.}
Scalar and categorical variables were treated as constant functions over the time domain,  facilitating their inclusion in the analysis without compromising feature selection {\orange consistency and showing \texttt{FSFC}'s proficiency in dealing with multi-modal data scenarios.}
{\red This adjustment results in \texttt{FSFC} estimating functional coefficients for these features rather than real numbers. These coefficients will be constant functions across the domain where the function's value aligns with the conventional logistic coefficient.}
We indicate the models for the selected categorical response variables as $m_{\text d}$ (diabetes), $m_{\text{inf}}$ (myocardial infarction), $m_{\text c}$ (high cholesterol), and $m_{\text{hyp}}$ (hypertension).

We explore two different experiment settings, detailed in Appendix Section \ref{sec:appendix_share_settings}. In \emph{Experiment 1}, we limit our analysis to a subset of 20 subjects to replicate a scenario where $p > n$. In \emph{Experiment 2},  we consider the complete data. In this case,  $n>p$.
For each response, in both experiments, all the methods were executed for 100 replications using the hyper-parameters detailed in the Appendix, Section \ref{sec:appendix_hyperparameters}.  


\paragraph{Classification results.} Figure \ref{fig:share_results} presents classification and selection results.  
In \emph{Experiment 1}, \texttt{FSFC} achieves the best median accuracy and boosts the performance of \texttt{SVM} and \texttt{LSTM}. 
In \emph{Experiment 2}, \texttt{SVM} and \texttt{FSFC} achieve superior and comparable outcomes. In this setting, with $n>p$, applying \texttt{FSFC} as a pre-processing step improves the effectiveness of \texttt{LSTM} and does not significantly impact \texttt{SVM}. Applying \texttt{FSFC} as a pre-processing step improves the effectiveness of \texttt{LSTM} -- which already avoids model over-fitting. Nonetheless, \texttt{FSFC} still has the unique capability of identifying crucial features and producing insightful and meaningful outcomes.

{\orange Appendix Table \ref{tab:fe_share_results} presents the accuracy results from integrating Feature Extraction with \texttt{LSTM} and \texttt{SVM}. In situations where the number of features is comparable to the sample size, Feature Extraction significantly boosts the performance of \texttt{SVM}, especially in the Experiment 2 settings. However, this approach does not identify the key features related to the binary response and does not preserve their longitudinal nature.}

\paragraph{Feature selection results.} 
It is worth noticing that the most frequently selected features by the four models are consistent across the two experiments, showing FSFC feature selection efficiency even when the number of samples is small. 
In addition, we compare FSFC responses with medical literature, finding evidence of the model accuracy in extracting correlations between features.
The four examined chronic diseases exhibit strong interconnectivity. Both $m_{\text {d}}$ and $m_{\text{inf}}$ identify hypertension and high cholesterol as crucial features, while $m_{\text {c}}$ and $m_{\text {hyp}}$ incorporate the remaining three investigated diseases in the model. 
Substantial medical literature supports the association between these diseases, including the following studies: \citet{kearney2005global, emerging2010diabetes, cosentino20202019, baigent2010efficacy}.
\emph{Gender} has been identified as significant in diabetes \citep{kautzky2016sex} and myocardial infarction (i.e. hearth attack) \citep{vaccarino1999sex, bairey2006insights}; 
\emph{age} is associated with myocardial infarction \citep{white1996age, avezum2005impact} and hypertension \citep{franklin1997hemodynamic, vasan2002residual}; 
\emph{bmi} is related to diabetes \citep{chan1994obesity, mokdad2003prevalence} and hypertension \citep{chobanian2003seventh, gelber2007prospective};
%
%
and the level of \emph{mobility} and \emph{recall test score} are factors relevant to myocardial infarction \citep{mora2007physical, haring2013cardiovascular}.
Model $m_{\text {c}}$ also selects \emph{drinking} behavior and quality of life factors such as \emph{income perception}, findings supported in \citet {rimm1999moderate, hare2014depression}.
Lastly, the number of doctor visits is selected as a relevant factor by all models, aligning with multiple studies that document the significant impact of chronic diseases on healthcare utilization \citep{hoffman1996persons, lehnert2011health}.

%
%
\section{Conclusions} 
\label{sec:discussion}
In this paper, we present an innovative method, \texttt{FSFC}, that jointly performs feature selection and classification of multivariate functional data.
We utilize the properties of Functional Principal Components and implement a novel variant of the DAL algorithm that leverages the sparsity structure of the dual Hessian information to significantly reduce the problem's dimensionality.
\texttt{FSFC}'s computational efficiency enables (i) handling high-dimensional scenarios where the number of features may far exceed the number of statistical units, (ii) performing an exhaustive search through the algorithm's hyperparameters. 

A simulation study demonstrates that \texttt{FSFC} outperforms \texttt{SVM} and \texttt{LSTM} in terms of classification accuracy and computational time {\orange when the number of features is much larger than the sample size. Moreover,} \texttt{FSFC}'s unique feature selection capability reduces problem dimensionality and enhances the efficacy of the competing methods.
By applying \texttt{FSFC} to data from the SHARE study, we identify well-documented relationships between four chronic diseases in the literature. Furthermore, \texttt{FSFC} uncovers other critical health and socio-demographic factors that play a significant role in differentiating between the healthy and affected groups, as supported by numerous research studies.

In future work, we plan to incorporate functional responses in the model. The response would no longer be a category but would describe a class-belonging probability that evolves over time. Additionally, one could consider more than two different classes as in the multinomial regression \citep{kwak2002multinomial}.
One limitation of the current \texttt{FSFC} implementation concerns the parameter $k$, i.e., the number of principal components used to approximate the functional features. Currently, $k$ must be the same for all features. Allowing different $k$ values would enable the use of more components for features with high variability without increasing the overall problem's dimension.

\section*{Acknowledgements}
The authors received funding from the European Union’s Horizon 2020 research and innovation program under grant agreement No. 945449.
The authors also thank their partners in IMEC for helping access the SHARE data.

%
%

\nocite{langley00}

\bibliography{bib}
\bibliographystyle{icml2024}

%
%
\newpage
\appendix
\onecolumn

\counterwithin{figure}{section}
\counterwithin{table}{section}
\renewcommand\thefigure{\thesection\arabic{figure}}
\renewcommand\thetable{\thesection\arabic{table}}

%
%
\section{Proofs}
\label{sec:appendix_proofs}


\subsection{Proof of Proposition 1}
\label{subsec:appendix_h_star}

Note that $h(XB)$ can be expressed as a separable sum, i.e.: $h(XB)= \sum_{i=1}^n h(X_{(i)}B)$, with $h(X_{(i)}B) = \log \big(1 + \exp \big( -Y_i \cdot (X_{(i)}B ) \big) \big)$, where $Y_i \in \{-1, 1\}$. Hence, we have \citep{boyd2004convex}:
\begin{equation*}
\label{eq:separable_sum}
	h^*(V) = \sum_{i=1}^n h^* (V_i) \ .
\end{equation*}
By definition, $h^*(V_i) = g\big(\tilde{b} \big)$, where $g(b) = b V_i - h(b)$ and $\tilde{b}_i=\max_{b}g(b)$. \\

\noindent
\underline{\emph{Case $Y_i = -1$, $V_i \in (0,1)$}}

We have $g(b) = b V_i - \log \left(1 + \text{e}^{b} \right)$.
To find $\tilde{b}$,
we derive $g(b)$ and set the derivative equal to 0, obtaining:
\begin{equation}
\label{eq:der_0}
    V_i =  \big( 1 + \text{e}^{\tilde{b}} \big)^{-1} \text{e}^{\tilde{b}} \ .  
\end{equation}
To solve for $\tilde{b}$, we take the logarithm of both side and note $ \log \big(1 + \text{e}^{\tilde{b}} \big) = h \big( \tilde{b} \big)$. Therefore,
\[
     \tilde{b} = \log(V_i) + h \big( \tilde{b} \big).
\]
To find an explicit form for $h \big( \tilde{b} \big)$, we manipulate \eqref{eq:der_0}:
\[
    \eqref{eq:der_0} 
    ~~\Leftrightarrow~~
    1 - V_i = 1 - \big( 1 + \text{e}^{\tilde{b}} \big)^{-1} \text{e}^{\tilde{b}}
    ~~\Leftrightarrow~~
    \log(1 - V_i) = -\log\big( 1 + \text{e}^{\tilde{b}} \big) \ .
\]
The last equality gives us $h \big( \tilde{b} \big) = -  \log(1 - V_i)$. Then, we can compute:
\[
   h^*(V_i) = g\big(\tilde{b} \big) = V_i \big(\log (V_i) - \log(1 - V_i) \big) + \log(1 - V_i) = V_i \log(V_i) + (1 - V_i) \log(1-V_i) \ .
\]

\vspace{0.3cm}
\noindent
\underline{\emph{Case $Y_i = 1$, $V_i \in (-1,0)$}}

We have $g(b) = b V_i - \log \left(1 + \text{e}^{_b} \right)$. Following the same steps of the previous case, we obtain:
\[
	h^*(V_i) = - V_i \log(-V_i) + (1 + V_i) \log(1+V_i) \ .
\]

\noindent
Considering $\abs{Y_1, V_1} < 1$ and combining the two cases, we finally obtain the desired result:
\[
	h^*(V_i) = (1 - \abs{Y_i V_i})\log (1 - \abs{Y_i V_i}) + \abs{Y_i V_i}\log (\abs{Y_i V_i}) \ .
\]

\subsection{Proof of Proposition 2}
\label{subsec:appendix_hessian}

A significant part of this proof relies on the results of \citet{boschi2023fasten}. The authors prove a similar proposition in the case of a scalar-on-function linear regression scenario where the same penalty $\pi$ is considered. \\

\noindent 
\textbf{\emph{(i)}}~~ We need to compute $\psi(V) = \mathcal{L}_\sigma (V\,|\,\Bar{Z},B)$.
Plugging $\Bar{Z}$ in Equation (5)
, we get: 
\begin{align}
    \label{eq:psi_medium}
    \begin{split}
        \psi(V) = h^*(V) + \pi^*(\bar Z) + \frac{1}{2 \sa} \sum_{j=1}^p \normbig{\prox_{\sa \pi}\big(T_{j} \big)}_2^2 - \frac{1}{2\sa} \sum_{j=1}^p\norm{B_{j}}_2^2 \ .
    \end{split}
\end{align}
Next, from \cite{boschi2023fasten}, we know: 
\begin{align*}
    \begin{split}
        \pi^* (\bar Z_{j}) = (\omega_j\la_2 / 2) \left((1 + \sa \omega_j \la_2)^{-1} \left(\norm{ T_{j}}_2 - \sa \omega_j \la_1 \right) \right)^2 \ .
    \end{split}
\end{align*}
Furthermore, starting from the definition of $\prox_{\sa \pi}$ given in Equation (8)
, we have: 
\begin{align*}
    \begin{split}
        \norm{\prox_{\sa \pi}\left(T_{j}\right)}_2^2 & = \left( (1 + \sa \omega_j \la_2)^{-1} \left(1- \norm{T_{j}}_2^{-1} \sa \omega_j \la_1 \right) \norm{T_{j}}_2 \right)^2  \\
        & = \left((1 + \sa \omega_j \la_2)^{-1} \left(\norm{ T_{j}}_2 - \sa \omega_j \la_1 \right) \right)^2 \ .
    \end{split}
\end{align*}
Therefore,
\[
	\pi^* (\bar Z_{j}) = (\omega_j\la_2 / 2) \norm{\prox_{\sa \pi}\left(T_{[j]}\right)}_2^2, 
\]
which leads to $\pi^* (\bar Z) = \sum_{j=1}^p (\omega_j\la_2 / 2) \norm{\prox_{\sa \pi}\left(T_{j}\right)}_2^2$. If we plug this expression of $\pi^* (\bar Z)$ in \eqref{eq:psi_medium}, we obtain the desired result. \\

\noindent 
\textbf{\emph{(ii)}}~~ Again, from \cite{boschi2023fasten}, we know: 
$$
	\nabla_V \left(\frac{1}{2\sa} \sum_{j=1}^p \left( 1 + \sa \omega_j \la_2 \right) \normbig{\prox_{\sa \pi}\big(T_{j}\big) }_2^2\right) = - X \prox_{\sa \pi}(T)~. 
$$
To complete the proof, it is sufficient to note that the derivative of $h^*(V_i) = (1 - \abs{Y_i V_i})\log (1 - \abs{Y_i V_i}) + \abs{Y_i V_i}\log (\abs{Y_i V_i})$ is:
$$
	\nabla h_i^*(V) = Y_i \log \big((1 - \abs{Y_iV_i}) ^{-1} \abs{Y_iV_i} \big) \ .
$$

\vspace{0.3cm}
\noindent 
\textbf{\emph{(iii)}}~~  From \citep{boschi2023fasten} we have :
$$
	\nabla_V \big(-X \prox_{\sa \pi} (T) \big) = \sa X_\mathcal{J} Q_\mathcal{J} X_\mathcal{J}^T \ .
$$
Note that $\nabla_i h^*(V)$ in (ii) depends only on $V_i$, which implies that all the off-diagonal elements of $H_{h^*}$ are 0. The $i$-th diagonal element of $H_{h^*}$ is then computed as follows:
$$
	H_{h^* (ii)} = \nabla_V \Big( Y_i \log \big((1 - \abs{Y_iV_i}) ^{-1} \abs{Y_iV_i} \big) \Big) = 
	\big( \abs{Y_iV_i}(1 - \abs{Y_iV_i})\big)^{-1} \ .
$$ 
Combining the two equations, we complete the proof.

%
%
\section{Algorithms' hyper-parameters}
\label{sec:appendix_hyperparameters}

For both \texttt{SVM} and \texttt{LSTM}, the inputs are the functional features evaluated on a grid of 100 uniformly spaced time points for the simulations scenarios, and 192 uniformly spaced time points (one for each month) for the SHARE application. 

The \texttt{SVM} method utilizes the default parameters from the \texttt{kernlab R package}, with \texttt{kernel = rbfdot}.

The \texttt{LSTM} architecture consists of two layers with 50 units each, activated by the \texttt{relu} function. This is followed by a dense layer that outputs class probabilities and is activated by the \texttt{softmax} function. Training is conducted over 200 epochs, utilizing batches of 32 and the \texttt{adam} optimizer.

In all scenarios, \texttt{FSFC} performs a path search on a grid of $100$ different $\lambda_1$ values. Specifically, $\lambda_1= c_\lambda\lambda_{max}$ and $\lambda_2=(1 - \alpha)\lambda_1$, where $\lambda_{max} = 0.5\max_j\lVert (X^TY)_{(j)}/\omega_j\rVert$, $~c_\lambda$ belongs to a grid of $100$ values spaced on a logarithmic scale from $1$ to $0.01$, and $\alpha$ is set to $0.2$.
After selecting the model that minimizes a 5-fold cross-validation criterion, the adaptive procedure  is implemented as detailed in the main text.
For each instance, \texttt{FSFC} starts from an initial value of $\sigma^0=0.1c_\lambda/\lambda_{max}$ increased by a factor of $\max(\min(5, 1 + 10c_\lambda), 1.1)$ at each iteration. Choosing $\sigma^0$ based on $\lambda_{max}$ and $c_\lambda$ is a common practice in the DAL optimization literature \citep{tomioka2011super}.
The tolerance in Equation (11),
employed for determining algorithm convergence, is set to $10^{-4}$, and the parameter $\mu$ in the line-search procedure in Equation (10)
is set to $0.2$.
Finally, we set $k=5$. Across all examined scenarios and for every feature, five functional principal components are enough to capture more than $95\%$ of the curves' variability, allowing for a good approximation of the features while maintaining manageable problem dimensions.
We explored various values of $\alpha$, $\mu$, and $k$, which yielded comparable outcomes not included in the paper.

%
%
\section{Synthetic data generation}
\label{sec:appendix_synthetic_data}

\begin{figure}[!h]
    \centering
    \includegraphics[width=1\textwidth]{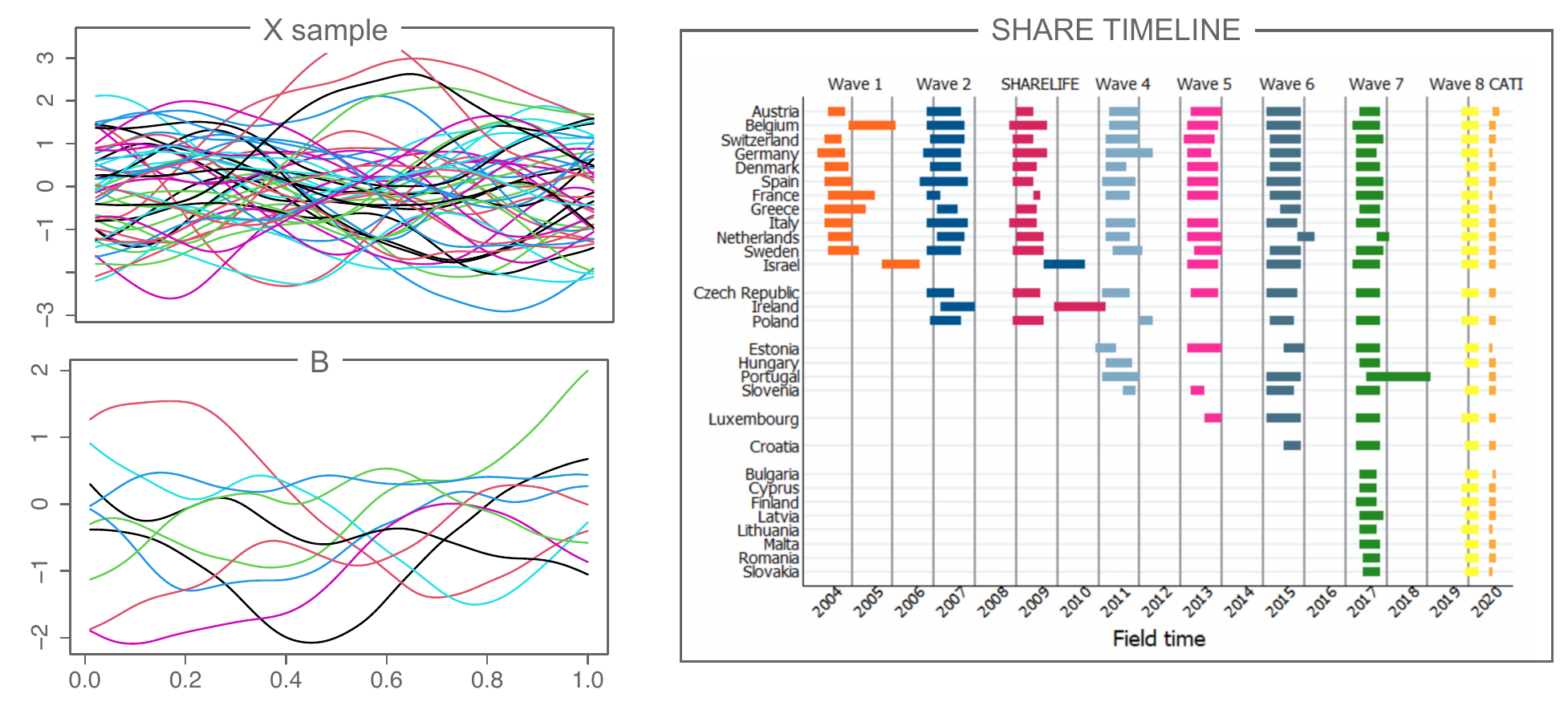}
    \vspace*{-0.5cm}
    \caption{On the left, a sample of 50 curves from the first feature of the design matrix $\mathcal{X}$ (top) and the 10 non-zero $\mathcal{B}$ coefficients (bottom) are displayed for the given scenario with $n=300$, $p=800$, $p_0=10$. On the right, the SHARE project timeline is depicted, which has been sourced from the SHARE website: \texttt{\href{https://share-eric.eu/data/data-documentation/waves-overview}{https://share-eric.eu/data/data-documentation/waves-overview/}}
    }
\label{fig:appendix}
\end{figure}

Following other work on functional regression 
\citep{parodi2018simultaneous}, we draw each functional feature $\mathcal {X}_{j}$ and each non-zero regression coefficient curve $\mathcal B_j$ from a zero-mean Gaussian process with a Matern covariance function \citep{cressie1999classes} of the form
\begin{equation*}
\small
\begin{split}
        C(t,s) = \frac{\eta^2}{\Gamma(\nu)2^{\nu-1}}\Bigg(\frac{\sqrt{2\nu}}{l} \lvert t-s \rvert\Bigg)^\nu K_\nu\Bigg(\frac{\sqrt{2\nu}}{l} \lvert t-s \rvert \Bigg) \ ,
\end{split}
\end{equation*}
where $K_\nu$ is a modified Bessel function. We employ point-wise variance $\eta^2=1$, range $l=0.25$, and smoothness parameter $\nu=3.5$ for both the $\mathcal X$'s and the $\mathcal B$'s.
Before running \texttt{FSFC}, we standardize each feature individually as $\left(\mathcal {X}_{j} - \text{mean}(t) \right) / \text{sd}(t)$, where $\text{mean}(t)$ and $\text{sd}(t)$ are the point-wise average and standard deviation of all instances computed at $t$.
Appendix Figure \ref{fig:appendix} shows some instances of $\mathcal X$ and $\mathcal B$ for one specific scenario.
The categorical response $Y$ is generated according to the logistic regression procedure: each $Y_i$ is drawn from a Bernoulli distribution on $\{-1, 1\}$ with probability $\hat p_i = 1/\big(1 +  \exp \big(X_{(i)} \hat B \big) \big)$.

%
%
\section{SHARE experiment settings}
\label{sec:appendix_share_settings}

In \emph{Experiment 2}, The cardinalities of the ``affected" and ``healthy" subject groups, denoted as $n_a$ and $n_h$, are as follows: $n_a = 89, 102, 313, 419$ and $n_h = 1235, 1160, 747, 577$ for diabetes, myocardial infarction, high cholesterol, and hypertension, respectively.
To prevent unbalanced scenarios, for each replication, we randomly select $n_a$ subjects from the healthy pool, resulting in a total of $n = 2n_a$ observations evenly balanced between the two classes. The observations are then divided into training and test sets, with $n_{test}= n/4$.

In \emph{Experiment 1}, each replication uses a training set of only 20 random subjects. All the other observations are part of the testing set.

%
%
\section{Additional figures and tables}
\label{sec:appendix_add_fig_tab}

\begin{table}[!h]
\caption{Simulation results for the integration of Feature Extraction with LSTM (\texttt{feLSTM}) and SVM (\texttt{feSVM}). 
For each curve, the following 12 attributes have been extracted: 
\emph{mean, variance, median, maximum, minimum, skewness, kurtosis, zero crossing rate, spectral entropy, spectral kurtosis, area under the curve,} and \emph{autocorrelation}.
The table reports the average Classification test accuracy and CPU processing time, measured in seconds, for 50 instances in each scenario.
}
\vspace{0.2cm}
\centerline{
\scalebox{0.85}{
\begin{tabular}{lrrrrr}
                                                                                                     &                                                   & \multicolumn{2}{c}{\textbf{Accuracy test}}                                                                           & \multicolumn{2}{c}{\textbf{CPU time}}                                                                               \\ \Xhline{2\arrayrulewidth}
\multicolumn{1}{l|}{}                                                                                & \multicolumn{1}{l|}{$p_0$}                        & \multicolumn{1}{l}{\texttt{feLSTM}} & \multicolumn{1}{l|}{\texttt{feSVM}} & \multicolumn{1}{l}{\texttt{feLSTM}} & \multicolumn{1}{l}{\texttt{feSVM}} \\ \Xhline{2\arrayrulewidth}
\multicolumn{1}{l|}{\multirow{4}{*}{\begin{tabular}[c]{@{}l@{}}$n = 300$\\ $p = 800$\end{tabular}}}  & \multicolumn{1}{r|}{\texttt{2}}  & 0.535                                                & \multicolumn{1}{r|}{0.495}                           & 67.7                                                 & 52.6                                                \\
\multicolumn{1}{l|}{}                                                                                & \multicolumn{1}{r|}{\texttt{5}}  & 0.532                                                & \multicolumn{1}{r|}{0.504}                           & 67.8                                                 & 52.9                                                \\
\multicolumn{1}{l|}{}                                                                                & \multicolumn{1}{r|}{\texttt{10}} & 0.538                                                & \multicolumn{1}{r|}{0.502}                           & 67.7                                                 & 52.6                                                \\
\multicolumn{1}{l|}{}                                                                                & \multicolumn{1}{r|}{\texttt{20}} & 0.539                                                & \multicolumn{1}{r|}{0.514}                           & 67.8                                                 & 52.6                                                \\ \hline \hline
\multicolumn{1}{l|}{\multirow{4}{*}{\begin{tabular}[c]{@{}l@{}}$n = 600$\\ $p = 2000$\end{tabular}}} & \multicolumn{1}{r|}{\texttt{2}}  & 0.526                                                & \multicolumn{1}{r|}{0.504}                           & 298.1                                                & 271.2                                               \\
\multicolumn{1}{l|}{}                                                                                & \multicolumn{1}{r|}{\texttt{5}}  & 0.526                                                & \multicolumn{1}{r|}{0.509}                           & 298.1                                                & 273.5                                               \\
\multicolumn{1}{l|}{}                                                                                & \multicolumn{1}{r|}{\texttt{10}} & 0.531                                                & \multicolumn{1}{r|}{0.504}                           & 298.2                                                & 271.2                                               \\
\multicolumn{1}{l|}{}                                                                                & \multicolumn{1}{r|}{\texttt{20}} & 0.530                                                & \multicolumn{1}{r|}{0.505}                           & 297.9                                                & 271.3 \\ \Xhline{2\arrayrulewidth}                                       
\end{tabular}}}
\label{tab:fe_sim_results}
\end{table}

\begin{table}[!h]
\caption{SHARE results for the integration of Feature Extraction with LSTM (\texttt{feLSTM}) and SVM (\texttt{feSVM}). For each curve, the extracted attributes are the ones listed in Table \ref{tab:fe_sim_results}. The table reports the average classification test accuracy for 100 instances for each response. 
}
\vspace{0.2cm}
\centerline{
\scalebox{0.85}{
\begin{tabular}{ll|rr}
                                                   &                 & \multicolumn{1}{l}{\texttt{feLSTM}} & \multicolumn{1}{l}{\texttt{feSVM}} \\ \Xhline{2\arrayrulewidth}
\multicolumn{1}{l|}{\multirow{4}{*}{Experiment 1}} & \texttt{Diabetes}        & 0.603                      & 0.731                      \\
\multicolumn{1}{l|}{}                              & \texttt{Heart Attack}    & 0.612                      & 0.732                      \\
\multicolumn{1}{l|}{}                              & \texttt{High Colesterol} & 0.556                      & 0.659                      \\
\multicolumn{1}{l|}{}                              & \texttt{Hypertension}   & 0.614                      & 0.703                      \\ \hline \hline
\multicolumn{1}{l|}{\multirow{4}{*}{Experiment 2}} & \texttt{Diabetes}        & 0.652                      & 0.979                      \\
\multicolumn{1}{l|}{}                              & \texttt{Heart Attack}    & 0.632                      & 0.974                      \\
\multicolumn{1}{l|}{}                              & \texttt{High Colesterol} & 0.596                      & 0.985                      \\
\multicolumn{1}{l|}{}                              & \texttt{Hypertension}   & 0.675                      & 0.987                      \\ \Xhline{2\arrayrulewidth}
\end{tabular}}}
\label{tab:fe_share_results}
\end{table}

\begin{table}[!h] 
  \caption{List of the variables analyzed within the SHARE application. 
  The letter adjacent to the variable name denotes whether it is \emph{longitudinal} (l), \emph{scalar} (s), or \emph{categorical} (c). The letter (a) denotes a scalar variable obtained taking an average across the waves where the values were available.
  For more detailed information, the reader should consult the SHARE project website: \texttt{\href{https://share-eric.eu/}{https://share-eric.eu/}}}
  \vspace{0.3cm}
  \centerline{
  \scalebox{1}{
  \begin{tabular}{ll}
  \toprule
    \textbf{variable} & \textbf{short description} \\
    \toprule
    \texttt{CASP (l)} & quality of life index  \\ 
    \texttt{doctor visits (l)} & number of doctor visits within the past year  \\
    \texttt{recall test (l)} & number of words recalled in the first trial  \\     
	\texttt{bmi (l)} & body mass index  \\     
	\texttt{adlwa (l)} & activities of daily living index \\
	\texttt{adla (l)} & sum of five daily activities \\
	\texttt{lgmuscle (l)} & large muscle index \\
	\texttt{mobilityind (l)} & mobility index \\
	\texttt{grossmotor (l)} & grossmotor skills index \\
	\texttt{finemotor (l)} & finemotor skills index \\
	\texttt{income perception (l)} & household able to make ends meet \\
	\texttt{eurod (l)} & depression index with EURO-D scale \\
	\texttt{income pct (l)} & household income percentiles \\
	\texttt{heart attack (l)} & 1 if the subject ever had the disease, 0 o.w.\\
	\texttt{high cholesterol (l)} & 1 if the subject ever had the disease, 0 o.w. \\
	\texttt{stroke (l)} & 1 if the subject ever had the disease, 0 o.w. \\
	\texttt{diabetes (l)} & 1 if the subject ever had the disease, 0 o.w. \\
	\texttt{COPD (l)} & 1 if the subject ever had the disease, 0 o.w. \\
	\texttt{cancer (l)} & 1 if the subject ever had the disease, 0 o.w. \\
	\texttt{ulcer (l)} & 1 if the subject ever had the disease, 0 o.w. \\
	\texttt{parkinson (l)} & 1 if the subject ever had the disease, 0 o.w. \\
	\texttt{cataracts (l)} & 1 if the subject ever had the disease, 0 o.w. \\
	\texttt{hip fracture (l)} & 1 if the subject ever had the disease, 0 o.w. \\
	\texttt{age (l)} & age of the subject \\
	\texttt{education years (s)} & years of education \\
	\texttt{number of children (a)} & number of children that are still alive   \\
	\texttt{numeracy test 1 (a)} & mathematical performance    \\
	\texttt{numeracy test 2 (a)} & mathematical performance  \\
	\texttt{drinking behavior (a)} & times a patient drunk in the last 6 months  \\
	\texttt{hospitalization days (a)} & days spent at the hospital in the last 6 months  \\
	\texttt{number of hospitalization (a)} & number of hospitalizations in the last 6 months  \\
	\texttt{gender (c)} & female or male \\
	\texttt{vaccinated (c)} &  being vaccinated during childhood\\
	\texttt{ever smoked (c)} & ever smoked daily \\
  \bottomrule
  \end{tabular}}}
  \label{tab:var_list} 
\end{table}

\end{document}